\documentclass[a4paper,fleqn]{cas-dc}

\usepackage[authoryear,longnamesfirst]{natbib}

\def\tsc#1{\csdef{#1}{\textsc{\lowercase{#1}}\xspace}}
\tsc{WGM}
\tsc{QE}
\tsc{EP}
\tsc{PMS}
\tsc{BEC}
\tsc{DE}

\begin{document}
\let\WriteBookmarks\relax
\def\floatpagepagefraction{1}
\def\textpagefraction{.001}

\title [mode = title]{Cross-pyramid consistency regularization for semi-supervised medical image segmentation}                    



\author[1]{Matúš Bojko}[]

\credit{Conceptualization of this study, Methodology, Software}

\affiliation[1]{organization={Faculty of Informatics and Information Technologies, Slovak University of Technology in Bratislava},
                addressline={Ilkovičova 6276/2}, 
                city={Bratislava},
                postcode={84216 }, 
                state={Slovakia}}

\author[1]{Maroš Kollár}[]

\author[1]{Marek Jakab}[]


\credit{Data curation, Writing - Original draft preparation}

\author[2]{Wanda Benesova}
[orcid=0000-0001-6929-9694]
\cormark[1]
\ead{vanda.benesova@fit.cvut.cz}

\affiliation[2]{organization={Czech Technical University in Prague, Faculty of Information Technology},
            addressline={Thakurova 9}, 
            city={Prague},
            postcode={160 00}, 
            country={Czech Republic}}

\cortext[cor1]{Corresponding author}


\begin{abstract}
Semi-supervised learning (SSL) enables training of powerful models with the assumption of limited carefully labelled data and a large amount of unlabeled data to support
the learning.
In this paper, we propose a hybrid consistency learning approach to effectively exploit unlabeled
data for semi-supervised medical image segmentation by leveraging Cross-Pyramid Consistency
Regularization (CPCR) between two decoders. First, we design a hybrid Dual Branch Pyramid
Network (DBPNet), consisting of an encoder and two decoders that differ slightly, each producing a pyramid
of perturbed auxiliary predictions across multiple resolution scales. Second, we present a learning
strategy for this network named CPCR that combines existing consistency learning and uncertainty
minimization approaches on the main output predictions of decoders with our novel regularization term.
More specifically, in this term, we extend the soft-labeling setting to pyramid predictions
across decoders to support knowledge distillation in deep hierarchical features. Experimental results
show that DBPNet with CPCR outperforms five state-of-the-art self-supervised learning methods and
has comparable performance with recent ones on a public benchmark dataset.

\end{abstract}


\begin{highlights}
\item We propose a Self-Supervised Learning method for medical image segmentation using two decoders, which includes:

\item 
Learning framework called Cross Pyramid Consistency Regularization (CPCR).
\item 
Hybrid Dual-Branch Pyramid Network for auxiliary pyramid predictions on multiple scales.
\item 
Learning of consistency and uncertainty minimization on the main decoder outputs.
\item 
A new regularization provides an additional performance boost.
\end{highlights}


\maketitle

\section{Introduction}
 The increasing number of available medical data, including 2D and 3D image modalities, that physicians need to interpret has fostered the development of tools and analysis algorithms for computer-aided diagnosis. At the same time, one of the needs and requirements of medical image analysis is the precise delineation of key structures and regions to support accurate diagnoses. Therefore, image segmentation plays a central role in medical imaging. In this field, segmentation aims to highlight anatomical structures and pathological changes in images. Their accurate identification can support essential tasks such as tumor extraction, tissue volume measurement, and various clinical applications, including intervention planning and disease diagnosis.  

Advancements in deep learning, particularly convolutional neural networks (CNNs) and recent Vision Transformers (ViTs) \cite{vit}, have enabled remarkable performance across a wide range of medical image segmentation tasks. However, fully supervised segmentation methods require large and well-annotated datasets for training. Acquiring dense and precise pixel or voxel-level annotations for medical images is both expensive and time-consuming for domain experts. As a result, methods that leverage unlabeled data, such as semi-supervised approaches, have gained significant research attention and become increasingly important. 

Semi-supervised learning (SSL) enables training of powerful models with the assumption of limited carefully labelled data and a large amount of unlabeled data to support the learning. In general, recent SSL methods for medical image segmentation are broadly categorized into two directions: \textit{pseudo-labeling} and \textit{consistency regularization}.

\subsection{Pseudo-label}
 Pseudo-label based methods utilize a form of self-training, where the model is initially trained on a limited set of annotated data and then used to generate pseudo-annotation masks for unlabeled data. These artificially generated annotations, along with their corresponding data, are incorporated into the labelled set, and training alternates between these phases. \cite{inf-net,effective-ct}. These methods mainly differ in the generation of pseudo-labels, and the main challenge is over-fitting on its noise. Other more recent approaches, such as \textit{cross-teaching} \cite{cross-teaching}, \textit{co-training} \cite{deep-co-training,dhc,redundant-co-training,ua-co-training} or \textit{cross-pseudo-supervision} \cite{cps,cor-aware-cps}, extend the self-training approach using multiple, commonly two individual networks that learn from each other's predictions and guide the training of their counterparts. 

\subsection{Consistency regularization}
 Consistency regularization approaches, alongside standard supervised loss, introduce a regularization term for unlabeled data based on the smoothness assumption, which states that small perturbations in the input should not significantly alter the model’s predictions. To effectively leverage unlabeled data, different perturbations (such as transformations) are applied to various views to reduce output discrepancies between them and maintain consistent predictions. This prediction enforcement is typically implemented on different levels: \textit{data-level} \cite{patch-shuffle,cut-paste,copy-paste}, \textit{task-level} \cite{dual-task} or \textit{network-level} consistency \cite{ua-mt,copy-paste,urpc,mcnet+,mtnet}. 

\textit{Network-level} consistency methods typically adopt a teacher-student structure to enforce similarity between the predictions of the student model and its momentum-based teacher model. Most notable approach is Mean Teacher \cite{mean-teacher}, while its extensions \cite{copy-paste,ua-mt} build upon it due to great performance. Unfortunately, alongside pseudo-labelling methods, they often require multiple models or multiple passes through the network, greatly increasing the number of parameters needed during training. As a result, single-model or multi-decoder architectures have garnered considerable attention. For example, Luo \textit{et al.} \cite{urpc} utilizes a single encoder-decoder network equipped with auxiliary segmentation heads to generate perturbed pyramid output predictions across multiple decoder scales. This framework leverages unlabeled data by measuring the uncertainty of scale-specific outputs and minimizing the discrepancies between each prediction and their mean, resulting in more accurate and robust segmentation results. MC-Net+ \cite{mcnet+} employs multiple decoders, each with a distinct up-sampling strategy to produce different probability outputs, which are then sharpened to serve as pseudo-labels for supervising one another. Building on this concept, MTNet \cite{mtnet} enhances the approach in two key ways: first, by using different attention mechanisms to generate more diverse decoder outputs, and second, by implementing robust consistency learning and knowledge distillation to handle noisy pseudo-labels using T-softmax and an additional average prediction-based uncertainty minimization, to increase the network’s overall confidence.

\textit{Data-level} consistency approaches require that model outputs remain consistent even when the input undergoes different random transformations or augmentations. For example, some recent works used \textit{patch-shuffled} transformations \cite{patch-shuffle}, \textit{cut-paste} augmentation \cite{cut-paste}, and \textit{copy-paste} augmentation \cite{copy-paste}. 
 
\textit{Task-level} consistency enforces consistency between different tasks rather than just relying on input perturbations. This approach leverages auxiliary or multi-task learning to extract additional information from the original data. In recent work \cite{dual-task}, dual-task consistency was enforced between pixel-level classification task and level set regression task.

\section{Contribution}
In this paper, we propose a hybrid network-level consistency learning method for efficient semi-supervised medical image segmentation. First, we propose a hybrid Dual Branch Pyramid Network (DBPNet), to not only obtain diverse predictions from main outputs of decoders (branches), but also inspired by Luo \textit{et al.} \cite{urpc}, obtain auxiliary pyramid predictions from different scales. In our case the multi-scale predictions are acquired from both decoder branches to incorporate it into dual-decoder scheme. Second, for this scheme we propose a Cross Pyramid Consistency Regularization (CPCR) strategy that leverages both labeled and unlabeled images for consistency learning and knowledge distillation. In CPCR, similar to MTNet\cite{mtnet}, each branch serves as a teacher of the other branch using soft label supervision between main outputs of decoders. In addition, CPCR applies this supervision on the auxiliary pyramid predictions between pairs of predictions across the adjacent scales of decoders. With this approach we demonstrate that incorporating consistency constraint into deep hierarchical features of double-decoder scheme can improve the model's learning capability and performance. On top of that, inspired by MTNet\cite{mtnet}, we apply uncertainty minimization-based regularization to average prediction across the main outputs of branches, to further improve network confidence.

The contribution of this work is two-fold: 1) A hybrid Dual Branch Pyramid Network (DBPNet) that produces auxiliary pyramid predictions from multiple scales of two decoders for training a SSL medical image segmentation method. 2) A learning framework for this network named Cross Pyramid Consistency Regularization (CPCR), that combines existing consistency learning and uncertainty minimization approaches applied on main outputs of decoders with our regularization term between auxiliary pyramid predictions across two decoders for additional performance boost.

\begin{figure*}  
    \centering
    \includegraphics[width=0.9\textwidth]{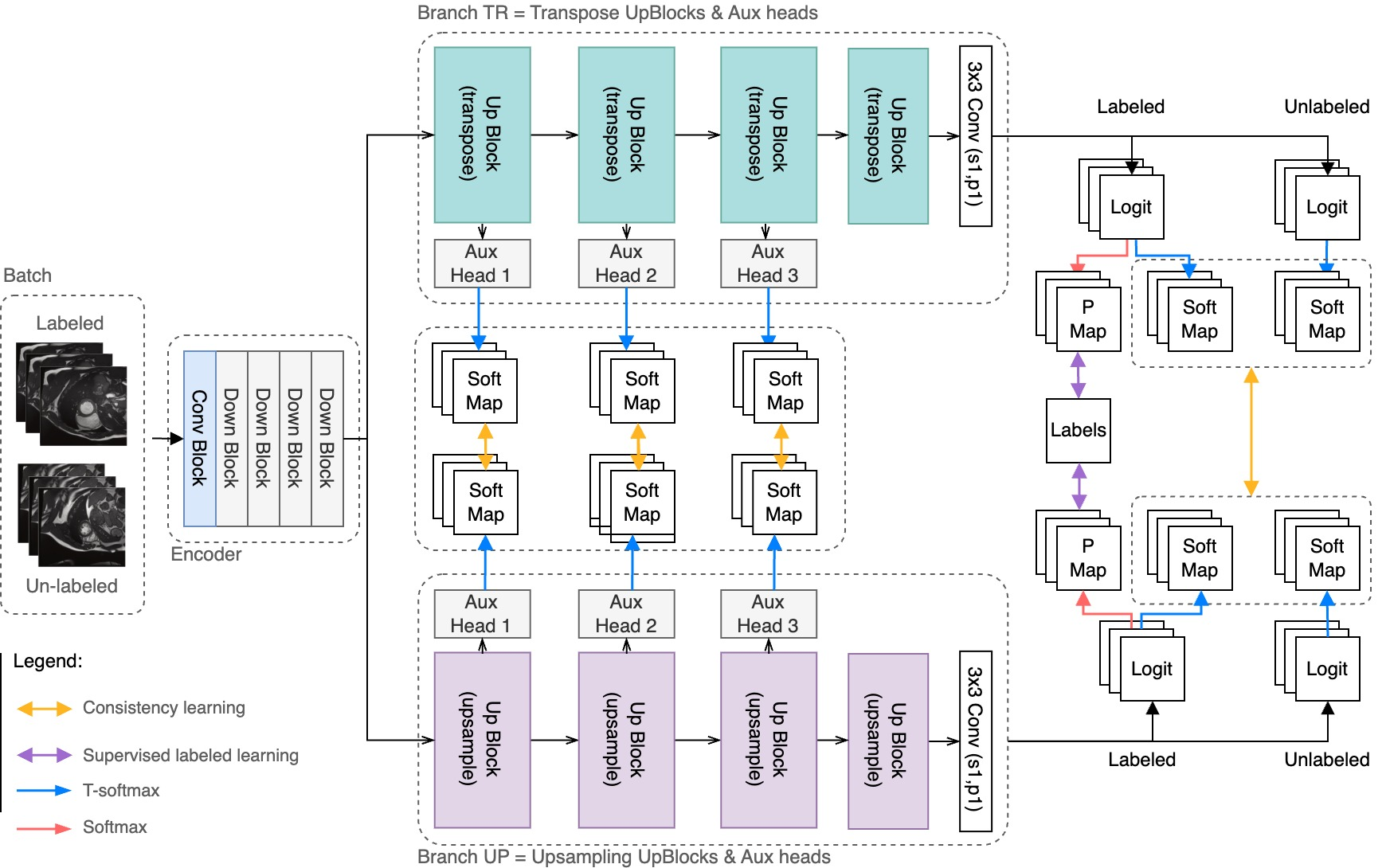}
    \caption{Overview of the proposed DBPNet and its learning framework via Cross Pyramid
Consistency Regularization (CPCR). Note that we did not depicted average prediction-based uncertainty minimization, as it is not the contribution of our work.}
    \label{proposal-image}
\end{figure*}

\section{Dual Branch Pyramid Network (DBPNet)}
Proposed \textbf{DBPNet} (in Fig.1) is based on U-Net \cite{u-net} structure with skip-connections, like most of other related methods, as it is still the main reference model for fair comparison of learning frameworks for SSL 2D medical image segmentation.
In contrast to MCNet+ \cite{mcnet+}, the multi-decoder network was simplified to two decoder branches, given that the third branch had a negligible impact on the final segmentation performance. We follow their strategy for obtaining output discrepancies by using dropout and different up-sampling method in each branch: transpose convolution or 1x1 convolution with bilinear up-sampling (denoted as \textit{TR} or \textit{UP} branch).

To produce the predictions at different pyramid scales, we added \textbf{auxiliary segmentation heads} from URPC \cite{urpc} after different resolution levels of both decoder branches (in total 3 scales) except the last main output (4th scale). 3x3 convolutional layer followed by upsampling and softmax was used to produce auxiliary predictions in input resolution. To produce output discrepancies at this level, every head is proceeded by different perturbation approach at each scale, either by applying: dropout, feature dropout or feature noise \cite{urpc}.

\subsection{Cross Pyramid Consistency Regularization}
Our proposed SSL framework consist of three regularization loss terms for consistency learning and uncertainty minimization. In single training iteration it uses both labeled and unlabeled data.

\paragraph{Main consistency loss}
 In the main loss term we enforce consistency and utilize knowledge distillation \cite{kd} between predictions from main outputs of each branch (on scale 4) using pseudo-label supervision. In this setting, the main prediction from one branch serves as pseudo-label for second branch and vice versa. To address noise present from incorrect predictions, we follow MTNet \cite{mtnet} by using KL divergence as distance metric and T-softmax for generating pseudo-labels as soft probability maps defined as:
\begin{equation} 
\tilde{\textbf{p}}_{c} = \frac{exp(\textbf{z}_c/T)}{\sum_{c}exp(\textbf{z}_c/T)}
\end{equation}
Where \textbf{z} is output logit map and T is a parameter to control the softness of the output probability. If \textit{TR} and \textit{UP} denotes branch, then the main loss term is:
\begin{equation}
        \mathcal{L}_{\text{con}}^{main} = KL(\tilde{P}_{TR}^{(4)},\tilde{P}_{UP}^{(4)}) + KL(\tilde{P}_{UP}^{(4)},\tilde{P}_{TR}^{(4)}) 
\end{equation}
Note that for the purpose of reciprocal supervision between branches in student-teacher manner, the KL divergence is applied in both directions (argument order) and gradient is back-propagated only into student branches. 

\paragraph{Auxiliary consistency loss}
 The auxiliary predictions up-sampled from different lower resolutions still contain different spacial frequencies i.e. they capture
the low-frequency component of the segmentation and contain low-level hierarchical features. \cite{urpc} We introduce an auxiliary consistency term to enforce consistency between these features across decoders. In this term we apply the soft label supervision between predictions at each scale of decoder branches to potentially support knowledge distillation across their deep layers. We defined it as average reciprocal supervision i.e:
\begin{equation}
        \mathcal{L}_{\text{con}}^{aux} = \frac{1}{3} \sum_{s=1}^{3} [KL(\tilde{P}_{TR}^{(s)},\tilde{P}_{UP}^{(s)}) + KL(\tilde{P}_{UP}^{(s)},\tilde{P}_{TR}^{(s)})]
\end{equation}
Where \textit{s} denotes the scale of the auxiliary prediction.

 As last regularization term we added \textbf{Average Prediction-based Uncertainty Minimization} from MTNet \cite{mtnet} to further encourage inter-decoder consistency through entropy minimization. It is defined as:
 \begin{equation}
    \mathcal{L}_{um} = - \frac{1}{N} \sum_{i=0}^{N} \sum_{c=0}^{C} \bar{P}_{i}^{c} log (\bar{P}_{i}^{c})
\end{equation}
With entropy minimization we can strength the model's confidence in its predictions. If we would apply it separately to each branch, it may lead to inconsistent prediction between them. That is why \(\bar{P} = ({P}_{TR}^{(s)} + {P}_{UP}^{(s)})/2\) is defined as average softmax probability map from main outputs of each branch (only at \(s=4\)). Then C and N are the class number and pixel number respectively.

For labeled data we calculate supervised loss term using Dice Loss between main output prediction of each branch (\({P}_{TR}^{(4)} ,{P}_{UP}^{(4)} \)) and ground truth label \(Y\).
\begin{equation}
        \mathcal{L}_{sup} = \mathcal{L}_{dice}(P_{TR}^{(4)},Y) + \mathcal{L}_{dice}(P_{UP}^{(4)},Y)
\end{equation}

At last, the overall loss function for our CPCR is weighted sum of all regularization terms and supervised loss, defined as:
 
\begin{equation}
    \mathcal{L}_{\text{total}} = \mathcal{L}_{\text{sup}} + 0.1 \cdot ( \mathcal{L}_{\text{con}}^{main} + \mathcal{L}_{\text{um}} ) +
\lambda \cdot \mathcal{L}_{\text{con}}^{aux} 
\end{equation}
Where \( \lambda(t)=w_{max} \cdot e^{(-5(1-\frac{t}{(t_{max}})^{2})}\) is time-dependent Gaussian
warming up function \cite{urpc} in a shape of sigmoid, to increase the weight during training (to \(w_{max}=0.1\) ), so that in early stages the model is not overwhelmed by incorrect predictions from deeper layers. Note that all regularization terms (\(\mathcal{L}_{\text{con}}^{main}, \mathcal{L}_{\text{con}}^{aux},\mathcal{L}_{\text{um}}\)) are applied on both labeled and unlabeled data. 

\section{Experiments and Results}
\subsection{Dataset}
For training and evaluation we utilized commonly used public benchmark dataset \textbf{ACDC (Automated Cardiac Diagnosis Challenge)} \cite{acdc-paper} , consisting of 200 annotated short-axis cardiac cine-MRI scans from 100 patients. Dataset is evenly divided into 5 subgroups according to condition:  normal (healthy), myocardial infarction, dilated cardiomyopathy, hypertrophic cardiomyopathy, abnormal right ventricle. All samples were collected from clinical examinations at the University Hospital of Dijon and obtained on 1.5T and 3T systems with typical resolution of 1.8×1.8×10.0mm3. For each patient, 2 scans were taken according to
different time window and cardiac status (end diastole or systole). Segmentation masks were manually annotated by their clinical expert and delineate 4 regions of interest (classes): left
ventricle endocardium (LV), left ventricle myocardium (Myo), right ventricle endocardium
(RV), and background. Due to large spacing (5mm) and possible shift between slices (due to respiration), the 3D scans were handled as multi-class segmentation task of the 2D short-axis slices. 
\subsection{Implementation details}
We implemented DBPNet and CPCR using PyTorch with help of MCNet+\cite{mcnet+} and public benchmark from Luo et al. \cite{urpc}. All dataset, hyper-parameter and augmentation settings (except num. of iterations) common with our method, including random seed (1337), followed the their works for fair comparison. 

In terms of data, we used the same fixed split of 70, 10 and 20 patients as training, validation and test sets respectively. At training time, the 2D slices were normalized to [0,1] and resized (zoomed) to 256x256 patches. To further extend the training set and avoid over-fitting, random flip and rotation were adopted as augmentations. During validation and testing time the 2D slices are resized to 256x256, fed into the network, enlarged back and stacked to originally sized 3D volumes.

We experimented with typical setup of 10\% labeled and the rest unlabeled data. This split, data sampling and augmentations were deterministic based on random seed (1337). During training we used SGD optimizer with momentum 0.9, weight decay 0.0001 and learning rate 0.01. The batch size was 24 with 12 labeled and 12 unlabeled data. Hyper-parameters were set to \(T=10, w_{max}, t_{max} = 200\) and we incremented \(t\) every 150 iteration. All our experiments were conducted on Azure Machine Learning with 6-core CPU, 112 RAM and V100 GPU for faster training time (around 6-12h). Out model was trained for 50k iterations, with validation phase every 200 iteration. For testing and validation we used our network only with \(UP\) branch, while the main (last) output (at scale 4) is used for final prediction. Following mentioned works \cite{mcnet+,urpc} Dice, Jaccard, the Average Surface Distance (ASD) and the 95\% Hausdorff Distance (95HD) were adopted for the quantitative evaluation.

\subsection{Comparison with existing works}
Our DBPNet with CPCR was compared with six existing SSL methods that also utilize consistency regularization in different ways. The experimental results for these methods (presented in Table \ref{results}) were obtained from Wu at al. \cite{mcnet+} and conducted under the same settings, which we largely followed to ensure a fair comparison. An exception is DVCPS \cite{dvcps} method which uses similar settings, but its results are sourced directly from its original paper. Note that displayed MCNet \cite{mcnet} is older variant of MTNet+ \cite{mcnet+}, but with two branches as our network.

To evaluate the effectiveness of semi-supervised learning, all methods were trained using only a 10\% annotation ration and compared against the standard U-Net \cite{u-net}, which serves as the backbone for all methods, including ours. The first two rows of the table display results for the fully supervised U-Net, using either 10\% or all available annotations. The displayed complexity shows number of parameters during inference.

Our DBPNet model with CPCR outperformed the related MCNet+ \cite{mcnet+} across all metrics, despite our model being architecturally smaller with two main branches (instead of three), which results in faster training due to a reduced number of parameters. However, DBPNet employs a more complex loss function, which may require careful tuning. Although we focused on optimizing DSC during prototyping with our method, DBPNet achieved the best performance on boundary-based metrics (95HD and ASD).

\begin{table}[htbp]
\centering
\scriptsize
\setlength{\tabcolsep}{2pt}
\begin{tabular}{l|cccc|c}
\hline
\textbf{Method:} & \multicolumn{4}{l|}{\textbf{Metrics:}} & \textbf{Complexity:} \\
& DSC(\%)$\uparrow$ & IoU(\%)$\uparrow$ & 95HD$\downarrow$ & ASD$\downarrow$ & Params.(M) \\
\hline
U-Net (10\%) & 77.34 & 66.20 & 9.18 & 2.45 & 1.81 \\
U-Net (All) & 91.65 & 84.93 & 1.89 & 0.56 & 1.81 \\
\hline
UA-MT (2019)  & 81.58 & 70.48 & 12.35 & 3.62 & 1.81 \\
SASSNet (2020) & 84.14 & 74.09 & 5.03 & 1.40 & 1.81 \\
DTC (2021) & 82.71 & 72.14 & 11.31 & 2.99 & 1.81 \\
URPC (2021) & 81.77 & 70.85 & 5.04 & 1.41 & 1.83 \\
MC-Net (2021) & 86.34 & 76.82 & 7.08 & 2.08 & 2.58 \\
MC-Net+ (2022)  & 87.10 & 78.06 & 6.68 & 2.00 & 1.81 \\
DVCPS (2025)  & \textbf{88.76} & \textbf{80.36} & 5.03 & 1.43 & -- \\
\textbf{DBPNet (Ours)} & 88.11 & 79.45 & \textbf{4.12} & \textbf{1.11} & 1.83 \\
\hline
\end{tabular}
\vspace{8pt}
\caption{ Comparison of proposed DBPNet and CPCR with six SSL methods on ACDC dataset, using 10\% of annotations for training}
\label{results}
\end{table}

\section{Conclusion}
We have presented a network-level consistency learning method for efficient semi-supervised medical image segmentation named Cross Pyramid Consistency Regularization (CPCR). This approach utilizes Dual Branch Pyramid Network (DBPNet) that produces auxiliary pyramid predictions from multiple scales of two decoders. Apart from main outputs of decoders, CPCR also utilizes soft label supervision between these auxiliary predictions at each scale of decoder branches to enforce consistency and potentially support knowledge distillation across their deep layers. Experimental results show that DBPNet with CPCR outperforms five state-of-the-art SSL methods and have comparable performance to the most recent ones on ACDC dataset with 10\% annotation ratio.

\section*{Acknowledgments}
This work was supported by the Slovak Research and Development Agency under the Contract no. APVV  SK-IL-RD-23-0004.

\newpage

\bibliographystyle{abbrv}
\bibliography{bibliography}

\vfill

\end{document}